\documentclass{article}

\usepackage[preprint]{neurips_2019}

\usepackage[utf8]{inputenc} % allow utf-8 input
\usepackage[T1]{fontenc}    % use 8-bit T1 fonts
\usepackage{hyperref}       % hyperlinks
\usepackage{url}            % simple URL typesetting
\usepackage{booktabs}       % professional-quality tables
\usepackage{amsfonts}       % blackboard math symbols
\usepackage{nicefrac}       % compact symbols for 1/2, etc.
\usepackage{microtype}      % microtypography

\usepackage{graphicx}
\usepackage[ruled,vlined]{algorithm2e}
\usepackage{booktabs}
\usepackage{amsmath}
\usepackage{float}

\usepackage{natbib}
\setcitestyle{authoryear,open={(},close={)}}
\hypersetup{colorlinks,linkcolor={blue},citecolor={blue},urlcolor={red}}

\title{Monocular Depth Estimation Using Multi Scale Neural Network And Feature Fusion}

\author{%
  Abhinav Sagar\thanks{Website of author - \url{https://abhinavsagar.github.io/}} \\
  Vellore Institute of Technology\\
  Vellore, Tamil Nadu, India\\
  \texttt{abhinavsagar4@gmail.com} \\
}

\begin{document}

\nocite{*}

\maketitle

\begin{abstract}
Depth estimation from monocular images is a challenging problem in computer vision. In this paper, we tackle this problem using a novel network architecture using multi scale feature fusion. Our network uses two different blocks, first which uses different filter sizes for convolution and merges all the individual feature maps. The second block uses dilated convolutions in place of fully connected layers thus reducing computations and increasing the receptive field. We present a new loss function for training the network which uses a depth regression term, SSIM loss term and a multinomial logistic loss term combined. We train and test our network on Make 3D dataset, NYU Depth V2 dataset and Kitti dataset using standard evaluation metrics for depth estimation comprised of RMSE loss and SILog loss. Our network outperforms previous state of the art methods with lesser parameters.

\end{abstract}

\section{Introduction}

Deep learning powered by neural networks has been successful in a range of problems in computer vision. Making autonomous Driving a reality requires solving the perception problem. There are a lot of sub-tasks involved like object detection, instance segmentation, depth estimation, scene understanding etc. Neural Networks tries to mimic the human brain by learning from the data without being explicitly programmed \citep{goodfellow2016deep}. In this work, we tackle the depth estimation problem especially in the context of autonomous driving. 

Depth estimation is an important but complex problem in computer vision. This requires learning a function which calculates the depth map from the input image. Humans have this ability naturally as their brain is able to understand the scene by making use of information from lighting, shading, perspective vision and presence of objects at various sizes \citep{godard2017unsupervised}. For humans it is pretty easy to infer the distance at which objects are present from a single image, however the task is quite challenging for a computer \citep{laina2016deeper}.

Stereo cameras have been traditionally used in Simultaneous Localization and Mapping (SLAM) based systems which has access to depth maps. However using monocular camera offers benefits like low power consumption, light weight and cheap. Hence this approach seems like a better alternative. In the literature, depth estimation has been mostly tackled using stereo cameras \citep{rajagopalan2004depth}. Depth estimation from a single image or monocular camera has been lately tackled using a range of convolutional network architectures \citep{eigen2014depth}, \citep{laina2016deeper} and \citep{liu2015learning}. The problem have been cast as a regression one which uses a Mean Square Error(MSE) in log space as the loss function.

\section{Related Work}

Early works on depth estimation were mostly based on stereo images using geometry based algorithms. Supervised learning was used to learn depth from monocular cues in 2D images \citep{saxena2008make3d}. A lot of work has been done using handcrafted techniques for feature extraction \citep{rajagopalan2004depth}. However these methods can only capture local information. Depth estimation has been tackled using image classification networks as feature extractors \citep{eigen2014depth} and \citep{garg2016unsupervised}. Spatial pyramidal pooling is used for reducing the spatial resolution of feature maps. 

Deep networks based on VGG and ResNet as feature extractors have been able to beat the previous techniques \citep{garg2016unsupervised} and \citep{eigen2014depth}. A multi scale network was used the low spatial resolution depth map to high spatial resolution \citep{eigen2014depth}. This helped reduce the recurring pooling operation which helped decrease the spatial resolution of feature maps. The network was divided into 2 parts: coarse network which predicts depth of the scene at a global level and fine network which uses local information to refine the depth.

Multi layer deconvolutional network has been used which uses high resolution feature maps \citep{laina2016deeper}. Residual upsampling modules were used along with a Resnet based feature extractor. \citep{jiao2018look} proposed a multi task convolutional neural network which uses lateral sharing approach between the individual networks. Also a new loss function was used to tackle the imbalanced depth distribution. \citep{fu2018deep} used a multi scale approach by discretizing the weights thus better taking uncertainty at various depths into account. VGG and Resnet based feature extractors were benchmarked along with atrous-spatial-pyramid-pooling approach to enhance the receptive field of the network. 

A skip connection based approach was used to fuse low spatial resolution depth map at deeper layers to high spatial resolution depth maps at previous layers \citep{xie2016deep3d}. To reduce the computational burden multi scale network has been used for extracting the features \citep{liu2015deep} and skip connections \citep{xie2016deep3d}. Also work has been done recently using unsupervised learning or semi supervised learning \citep{garg2016unsupervised}. Reconstruction losses are used for estimating the disparity map by using information from both the left and right view.

Our main contributions can be summarized as:

• We propose a novel end to end trainable network for monocular depth estimation.

• We present the network architecture, training details, loss functions and ablation studies.

• Our network outperforms previous state of the art networks on Make3D Range Image Data, NYU Depth Dataset V2 and Kitti dataset.

\section{Method}

\subsection{Dataset}

The following datasets have been used for training and testing our network:

1. \textbf{Make3D Range Image Data} - This dataset was one of the first proposed to infer the depth map from a single image. It has the range data corresponding to each image. Examples from the dataset include outdoor scenes, indoor scenes and synthetic objects \citep{saxena2008make3d}.

2. \textbf{NYU Depth Dataset V2} - This dataset is made up of video sequences from a variety of indoor scenes which have been recorded using both RGB and depth cameras. It has 1449 densely labeled pairs of aligned RGB and depth images. The objects present in the dataset have been individually labelled with a class id \citep{silberman2012indoor}. The official split consists of 249 training and 215 testing scenes. The images are of resolution is 480$\times$640. 

3. \textbf{Kitti dataset} - This large dataset has over 93 thousand depth maps with corresponding raw Lidar scans and RGB images. This has been the benchmark dataset for depth estimation using a single image for autonomous driving \citep{geiger2013vision}. For benchmarking, Eigen split was done by \citep{eigen2014depth}. The training set consists of
approximately 22 600 frames from a total of 28 different scenes and the validation set contains of 888 frames. The test set contains 697 frames from 28 different scenes. The images are of resolution 376$\times$1242.

\subsection{Data Augmentation} 

Data Augmentation is the process in which the dataset size is manually increased by performing operations on the individual samples of the dataset. This leads to better generalization ability thus avoiding overfitting of the network. Data Augmentation has been used successfully for depth estimation \citep{alhashim2018high} and \citep{li2018monocular}.

The training data was increased using data augmentation:

• \textbf{Scale}: Colour images are scaled by a random number
$s \in [1, 1.5]$.

• \textbf{Rotation}: The colour and depth images are both rotated with a random degree $r \in [-5, 5]$.

• \textbf{Colour Jitter}: The brightness, contrast, and saturation of
color images are each scaled by $k \in [0.6, 1.4]$.

• \textbf{Colour Normalization}: RGB images are normalized through mean subtraction and division by standard deviation.

• \textbf{Flips}: Colour and depth images are both horizontally flipped with a 50\% chance.

Also nearest neighbor interpolation was used.

\subsection{Network Architecture}

The task is to learn a direct mapping from a colour image to the corresponding depth map. Our network fuses multi scale depth features which is important for depth estimation. Our network removed all the fully connected layers which adds a lot of computational overhead. Although fully connected layers are important in inferring long range contextual information but still it is not required. Instead we use dilated convolutions which enlarges the receptive field without increasing the number of parameters involved.  

The network takes as input an image and uses a pre trained ResNet backbone for feature extraction. Convolutions are used at multiple scales with combinations of 1$\times$1 convolution, 3$\times$3 convolution, 5$\times$5 convolution and 7$\times$7 convolution. Instance-wise concat operation is performed to merge the feature maps. This multi scale block is repeated for 4 times. The receptive field of our network increases considerably due to this operation and is able to capture global contextual information in addition to the local information. 

The fused features is propagated to another multi scale block. This block is made up of plain convolutional layer and dilated convolutions with dilation rates of 2 and 4 respectively. This block is also repeated for 4 times and instance-wise concat operation is used for merging the feature maps. The network architecture used in this work is presented in Figure 1: 

\begin{figure}[htp]
    \centering
    \includegraphics[width=14cm]{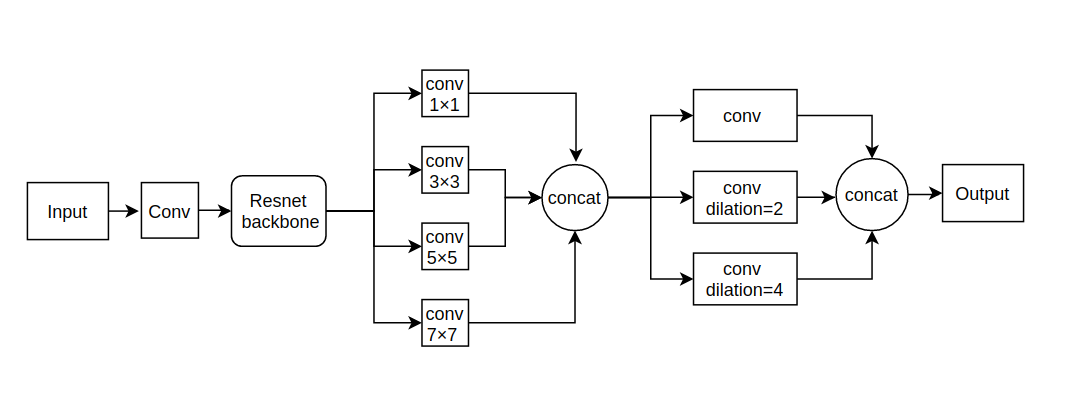}
    \caption{Network architecture used in this work}
    \label{fig1}
\end{figure}

\subsection{Multi Scale Fusion}

The high level neurons have a larger receptive field in convolutional neural network. Although low level neurons has a smaller receptive field, it contains more detailed information. Hence for better results, we combined feature maps at different scales. We concatenated the high level and intermediate level feature maps using a concat operator. Skip connections also helps the multi scale fusion operation by creating an additional pathway for flow of information.

\subsection{Loss Functions}

The standard loss function for training depth estimation network is a regression loss which is the difference between the ground-truth depth map $y$ and the prediction of the network $\hat{y}$ \citep{eigen2014depth}. Loss functions are very important for avoiding training instability as well as achieving better results. A lot of loss functions have been proposed in literature for depth estimation \citep{fu2018deep} and \citep{laina2016deeper}. We design our loss function by minimizing the reconstruction depth and penalizing the high frequency details. Our loss function is made up of 3 terms: depth term, Structural Similarity Index Measure (SSIM term) and a multinomial logistic loss term. The depth term is a L1 loss defined on the depth values as shown in Equation 1:

\begin{equation}
L_{d e p t h}(y, \hat{y})=\frac{1}{n} \sum_{p}^{n}\left|y_{p}-\hat{y}_{p}\right|
\end{equation}

SSIM metric is frequently used for measuring the image quality and similarity between two images. \citep{godard2017unsupervised} first used this while training depth estimation network. The upper bound of SSIM metric is 1, hence the loss term can be defined as in Equation 2:

\begin{equation}
L_{S S I M}(y, \hat{y})=\frac{1-S S I M(y, \hat{y})}{2}
\end{equation}

We cast depth estimation task as a kind of image classification one. Multinomial
logistic loss term is defined as in Equation 3:

\begin{equation}
L(\theta)=-\left[\sum_{i=1}^{N} \sum_{k=1}^{K} \left\{y^{(i)}\right\} \log \frac{\exp \left(\theta^{(k) T} y^{(i)}\right)}{\sum_{i=1}^{K} \exp \left(\theta^{(i) T} y^{(i)}\right)}\right]
\end{equation}

where $N$ is the number of training samples, $exp(\theta(k)T x(i))$ is the probability of label $k$ of sample $i$, and $k$ is the ground truth label.

The three terms can be combined together to yield the complete loss function which is used to train the network as in Equation 4:

\begin{equation}
L(y, \hat{y})=\alpha L_{\text {depth}}(y, \hat{y})+\beta L_{S S I M}(y, \hat{y})+ \gamma L(\theta)
\end{equation}

Where $\alpha$, $\beta$ and $\gamma$ are constants.

\subsection{Evaluation Metrics}

For evaluating depth predicting networks, the error metrics used by \citep{eigen2014depth} are commonly used. Let $y_{i}$ denotes the prediction value of pixel, $y_{i}^{\star}$ the ground truth value of pixel $i$ and $T$ denotes the total number of pixels which are valid. The error metrics are defined in the form of Root Mean Square Error (RMSE) and RMSE($\log$) as defined in Equation 5 and Equation 6 respectively:

\begin{equation}
\mathrm{RMSE}=\sqrt{\frac{1}{T} \sum_{i}\left\|y_{i}-y_{i}^{*}\right\|^{2}}
\end{equation}

\begin{equation}
\operatorname{RMSE}(\log )=\sqrt{\frac{1}{T} \sum_{i}\left\|\log \left(y_{i}\right)-\log \left(y_{i}^{*}\right)\right\|^{2}}
\end{equation}

The SILog error metric was defined by \citep{eigen2014depth} to measure the relationship between points in the scene irrespective of the absolute global scale which is shown in Equation 8. The value of $d_{i}$ can be computed using Equation 7:

\begin{equation}
d_{i}=\log \left(y_{i}\right)-\log \left(y_{i}^{*}\right)
\end{equation}
 
\begin{equation}
\operatorname{SILog}=\frac{1}{T} \sum_{i} d_{i}^{2}-\frac{1}{T^{2}}\left(\sum_{i} d_{i}\right)^{2}
\end{equation}

The Averaged Relative Error (ARE) and Squared Relative Error (SRE) metrics is defined in Equation 9 and Equation 10 respectively:

\begin{equation}
\mathrm{ARE}=\sqrt{\frac{1}{T} \sum_{i} \frac{\left|y_{i}-y_{i}^{*}\right|}{y_{i}^{*}}}
\end{equation}

\begin{equation}
\mathrm{SRE}=\sqrt{\frac{1}{T} \sum_{i}} \frac{\left\|y_{i}-y_{i}^{*}\right\|^{2}}{y_{i}^{*}}
\end{equation}

Accuracy with a threshold metric, divides the error ratios 
into intervals determined by the threshold value $\lambda$. The accuracy is defined as the number of pixels with a error ratio less than the threshold divided by the
total number of pixels present. This error metric is shown in Equation 11:

\begin{equation}
\frac{1}{T} \sum_{i}\left(\max \left(\frac{y_{i}}{y_{i}^{*}}, \frac{y_{i}^{*}}{y_{i}}\right)=\delta<\mathrm{thr}\right), \mathrm{thr}=\left[\lambda, \lambda^{2}, \lambda^{3}\right]
\end{equation}

The value of $\lambda$ is taken as 1.25. 

For quantitative evaluation, error metrics Mean relative error and Mean $\log _{10}$ error is defined in Equation 12 and Equation 13 respectively:

\begin{equation}
\text { Rel } = \frac{1}{|T|} \sum_{d \in T}|\hat{d}-d| / d
\end{equation}

\begin{equation}
\log _{10} = \frac{1}{T \mid} \sum_{d \in T}\left|\log _{10} \hat{d}-\log _{10} d\right|
\end{equation}

Where $d$ represents the ground truth depth, $\hat{d}$ represents the estimated depth, and $T$ denotes the
set of all points in the images.

\subsection{Implementation Details}

State of the art ResNet backbone was used as feature extractor which is trained on the Imagenet dataset. In all the experiments, ADAM optimizer was used with a learning rate value of 0.0001, parameter values momentum as 0.9, weight decay value of 0.0004 and batch size is set to 8. The network was trained using Stochastic Gradient Decent (SGD) for 500K iterations for NYU Depth v2 dataset, 100K iterations for Make3D dataset and 300K iterations for Kitti dataset.

\section{Results}

The comparison of our network with previous state of the art methods on NYU Depth v2 dataset is shown in Table 1: 

\begin{table}[hbt!]
  \caption{Performance on NYU Depth v2 dataset. 2nd, 3rd and 4th column: higher is better; 5th, 6th and 7th column: lower is better.}
  \label{h3}
  \centering
  \begin{tabular}{lllllll}
  \toprule
    Method &$\delta_{1}$ &$\delta_{2}$ &$\delta_{3}$ &rel &$log_{10}$ &rms\\
    \midrule
\citep{saxena2008make3d} &0.447 &0.745 &0.897 &0.349 &- &1.214\\
\citep{karsch2014depth} &- &- &- &0.35 &0.131 &1.2\\
\citep{liu2010single} &- &- &- &0.335 &0.127 &1.06\\
\citep{li2018monocular} &0.621 &0.886 &0.968 &0.232 &0.094 &0.821\\
\citep{wang2015towards} &0.605 &0.890 &0.970 &0.220 &- &0.824\\
\citep{roy2016monocular} &- &- &- &0.187 &- &0.744\\
\citep{liu2010single} &0.650 &0.906 &0.976 &0.213 &0.087 &0.759\\
\citep{eigen2014depth} &0.769 &0.950 &0.988 &0.158 &- &0.641\\
\citep{laina2016deeper} &0.629 &0.889 &0.971 &0.194 &0.083 &0.790\\
\citep{xu2017multi} &0.811 &0.954 &0.987 &0.121 &0.052 &0.586\\
\citep{fu2018deep} &0.828 &0.965 &0.992 &0.115 &0.051 &0.509\\
Ours &0.823 &0.962 &0.994 &0.101 &0.054 &0.456\\
    \bottomrule
  \end{tabular}
\end{table}

The model predictions compared along with ground truth depth map on NYU v2 dataset is shown in Figure 2:

\begin{figure}[H]
    \centering
    \includegraphics[width=12cm]{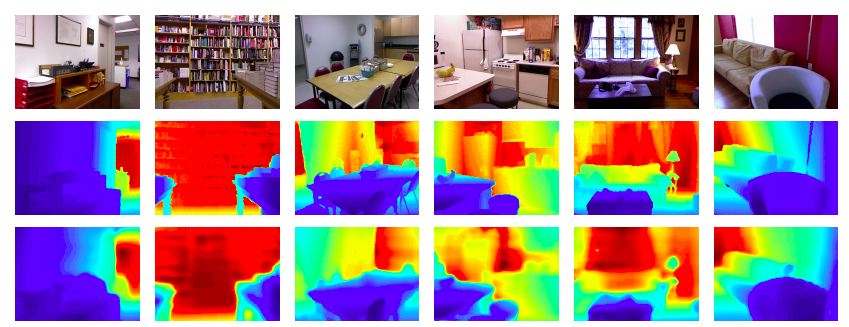}
    \caption{Qualitative comparison of the estimated depth map on the NYU v2 dataset. Color indicates depth (red is far, blue is
close). First row: RGB image, second row: Ground Truth depth map, third row: Results of our proposed method}
    \label{fig4}
\end{figure}

The comparison of our network with previous state of the art methods on Kitti dataset is shown in Table 2: 

\begin{table}[hbt!]
  \caption{Performance on KITTI dataset. All the methods are evaluated on the test split by \citep{eigen2014depth}. 3rd, 4th and 5th column: higher is better; 6th, 7th, 8th and 9th column: lower is better.}
  \label{h1}
  \centering
  \begin{tabular}{lllllllll}
  \toprule
    Method &cap &$\delta < 1.25$ &$\delta < 1.25^{2}$&
$\delta < 1.25^{3}$ &Abs Rel &Sq Rel &RMSE &$RMSE_{\log}$\\
    \midrule
\citep{saxena2008make3d} &0 - 80 m &0.601 &0.820 &0.926 &0.280 &3.012& 8.734 &0.361\\
\citep{eigen2014depth} &0 - 80 m &0.692 &0.899 &0.967 &0.190 &1.515 &7.156 &0.270\\
\citep{liu2010single} &0 - 80 m &0.647 &0.882 &0.961 &0.217 &1.841 &6.986 &0.289\\\
\citep{godard2017unsupervised} &0 - 80 m &0.861 &0.949 &0.976 &0.114& 0.898 &4.935 &0.206\\
\citep{kuznietsov2017semi} &0 - 80 m &0.862 &0.960 &0.986 &0.113 &0.741 &4.621 &0.189\\
\citep{fu2018deep} &0 - 80 m &0.915 &0.980 &0.993 &0.081 &0.376& 3.056 &0.132\\
\citep{fu2018deep} &0 - 80 m &0.932 &0.984 &0.994 &0.072 &0.307& 2.727 &0.120\\
\midrule
\citep{garg2016unsupervised} &0 - 50 m &0.740 &0.904 &0.962 &0.169 &1.080 &5.104 &0.273\\
\citep{godard2017unsupervised} &0 - 50 m &0.873 &0.954 &0.979 &0.108& 0.657 &3.729 &0.194\\
\citep{kuznietsov2017semi} &0 - 50 m &0.875 &0.964 &0.988 &0.108 &0.595 &3.518 &0.179\\
\citep{fu2018deep} &0 - 50 m &0.920 &0.982 &0.994 &0.079 &0.324& 2.517 &0.128\\
\citep{fu2018deep} &0 - 50 m &0.936 &0.985 &0.995 &0.071 &0.268& 2.271 &0.116\\
Ours &0 - 50 m &0.945 &0.987 &0.997 &0.066 &0.268& 2.042 &0.110\\
    \bottomrule
  \end{tabular}
\end{table}

The model predictions compared along with ground truth depth map on test image number 1 on Kitti dataset is shown in Figure 3:

\begin{figure}[H]
    \centering
    \includegraphics[width=9cm]{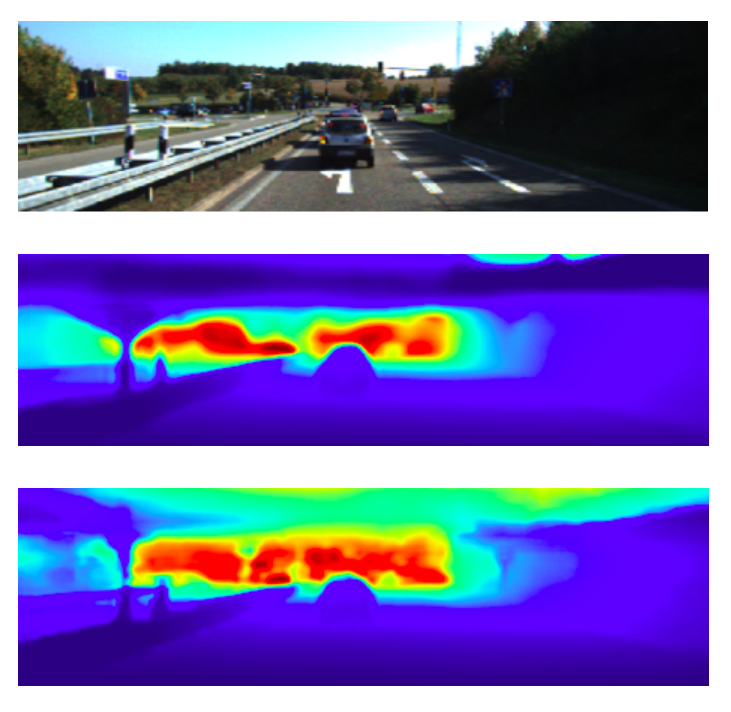}
    \caption{The output predictions of our network on
test image number 1. First row: input image, second row: ground truth depth map, third row: model prediction depth map. Color indicates depth (red is far, blue is
close).}
    \label{fig2}
\end{figure}

The model predictions compared along with ground truth depth map on test image number 5 on Kitti dataset is shown in Figure 4:

\begin{figure}[H]
    \centering
    \includegraphics[width=9cm]{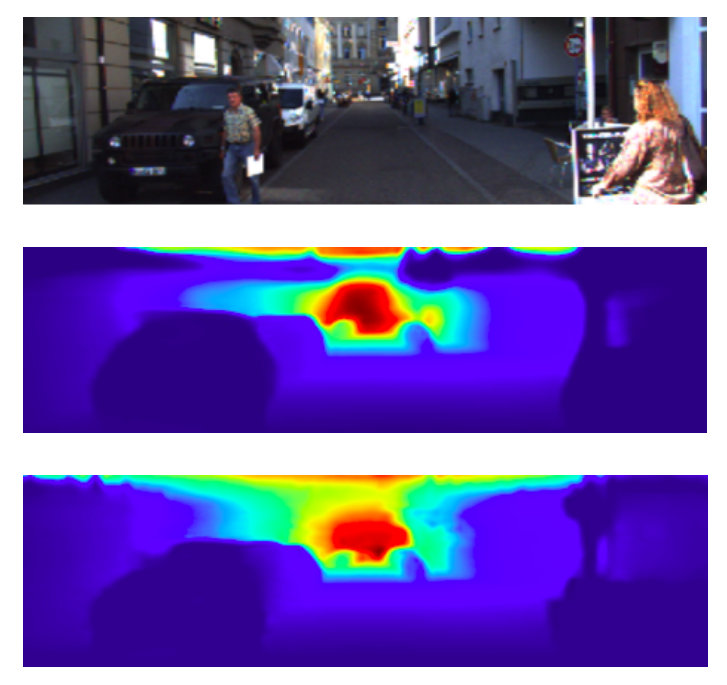}
    \caption{The output predictions of our network on
test image number 5. First row: input image, second row: ground truth depth map, third row: model prediction depth map. Color indicates depth (red is far, blue is
close). Our network fails to detect person in front of the car as well as the person in the bottom left corner}
    \label{fig3}
\end{figure}

The comparison of our network with previous state of the art methods on Make3D dataset is shown in Table 3: 

\begin{table}[hbt!]
  \caption{Performance on Make3D dataset. 2nd, 3rd and 4th column represents C1 error; 5th, 6th and 7th column represents C2 error. Both lower C1 and C2 error is better.}
  \label{h2}
  \centering
  \begin{tabular}{lllllll}
  \toprule
    Method &rel &$\log_{10}$ &rms &rel &$\log_{10}$ &rms\\
    \midrule
\citep{saxena2008make3d} &- &- &- &0.370 &0.187 &-\\
\citep{liu2010single} &- &- &- &0.379 &0.148 &-\\
\citep{karsch2014depth} &0.355 &0.127 &9.20 &0.361 &0.148 &15.10\\
\citep{liu2014discrete} &0.335 &0.137 &9.49 &0.338 &0.134 &12.60\\
\citep{liu2015deep} &0.278 &0.092 &7.12 &0.279 &0.102 &10.27\\
\citep{liu2015learning} &0.287 &0.109 &7.36 &0.287 &0.122 &14.09\\
\citep{roy2016monocular} &- &- &- &0.260 &0.119 &12.40\\
\citep{laina2016deeper} &0.176 &0.072 &4.46 &- &- &-\\
\citep{xie2016deep3d} &1.000 &2.527 &19.11 &- &- &-\\
\citep{godard2017unsupervised} &0.443 &0.156 &11.513 &- &- &-\\
\citep{kuznietsov2017semi} &0.421 &0.190 &8.24 &- &- &-\\
\citep{xu2018structured} &0.184 &0.065 &4.38 &0.198 &- &8.56\\
\citep{fu2018deep} &0.236 &0.082 &7.02 &0.238 &0.087 &10.01\\
\citep{fu2018deep} &0.157 &0.062 &3.97 &0.162 &0.067 &7.32\\
Ours &0.139 &0.060 &2.64 &0.144 &0.059 &6.36\\
    \bottomrule
  \end{tabular}
\end{table}

\subsection{Ablation Studies}

We perform ablation studies to analyze the performance of our network. The comparative performance using dilation and concat layers is shown in Table 4:

\begin{table}[hbt!]
  \caption{Ablation Study of our CNN architecture design on Kitti dataset. 2nd, 3rd and 4th column: higher is better; 5th, 6th and 7th column: lower is better.}
  \label{h6}
  \centering
  \begin{tabular}{lllllll}
  \toprule
    Method &$\delta < 1.25$ (\%) &$\delta < 1.25^{2}$ (\%) &$\delta < 1.25^{3}$ (\%) &Rel &$\log_{10}$ &rms\\
    \midrule
    no dilation no cocat &76.06 &94.29 &97.56 &0.156 &0.056 &0.536\\
no dilation yes concat &79.24 &96.2 &97.80 &0.145 &0.056 &0.520\\
yes dilation no concat &81.52 &95.43 &98.63 &0.132 &0.060& 0.533\\
yes dilation yes concat &83.10 &95.3 &98.70 &0.134 &0.051 &0.515\\
    \bottomrule
  \end{tabular}
\end{table}

\section{Conclusions}

In this paper, we proposed a novel network architecture for monocular depth estimation using multi scale feature fusion. We present the network architecture, training details, loss functions and the evaluation metrics used. We used Make 3D dataset, NYU Depth V2 dataset and Kitti dataset for training and testing our network. Our network not only beats the previous state of the art methods on monocular depth estimation but also has lesser parameters thus making it feasible in a real time setting.

\subsubsection*{Acknowledgments}

We would like to thank Nvidia for providing the GPUs for this work.

\bibliography{neurips_2019}

\end{document}